\documentclass[runningheads]{llncs}
\usepackage[T1]{fontenc}
\usepackage{graphicx}

\usepackage{hyperref}
\usepackage{color}

\usepackage{multirow}
\usepackage{amsmath}
\usepackage{amsfonts}
\usepackage{amssymb}
\usepackage{xspace}
\makeatletter
\DeclareRobustCommand\onedot{\futurelet\@let@token\@onedot}
\def\@onedot{\ifx\@let@token.\else.\null\fi\xspace}
 
\def\ie{\emph{i.e}\onedot}

\def\etal{\emph{et al}\onedot}
\makeatother

\usepackage{caption}
\usepackage{subcaption}

\begin{document}

\title{MSDA: Monocular Self-supervised Domain Adaptation for 6D Object Pose Estimation}

\titlerunning{MSDA}


\author{Dingding Cai\inst{1} \and
Janne Heikkilä\inst{2} \and
Esa Rahtu\inst{1}}
\authorrunning{D. Cai et al.}
\institute{
Tampere University, Finland \\
\email{\{dingding.cai, esa.rahtu\}@tuni.fi} \\
\and University of Oulu, Finland  \\
\email{janne.heikkila@oulu.fi}
}

\maketitle              
\begin{abstract}

Acquiring labeled 6D poses from real images is an expensive and time-consuming task. Though massive amounts of synthetic RGB images are easy to obtain, the models trained on them suffer from noticeable performance degradation due to the synthetic-to-real domain gap. To mitigate this degradation, we propose a practical self-supervised domain adaptation approach that takes advantage of real RGB(-D) data without needing real pose labels. We first pre-train the model with synthetic RGB images and then utilize real RGB(-D) images to fine-tune the pre-trained model. The fine-tuning process is self-supervised by the RGB-based pose-aware consistency and the depth-guided object distance pseudo-label, which does not require the time-consuming online differentiable rendering. We build our domain adaptation method based on the recent pose estimator SC6D and evaluate it on the YCB-Video dataset. We experimentally demonstrate that our method achieves comparable performance against its fully-supervised counterpart while outperforming existing state-of-the-art approaches.

\end{abstract}

\section{Introduction}
Estimating the 6D object pose from a monocular RGB image is one of the fundamental tasks in computer vision. This task aims to infer the 3D rotation and 3D translation of a rigid object with respect to the camera coordinate system, which is an important application in robotic manipulation \cite{collet2011moped,tremblay2018deep} and augmented reality \cite{marchand2015pose}. Due to the success of deep learning on large-scale image classification tasks \cite{russakovsky2015imagenet}, learning-based 6D pose estimation methods \cite{di2021so,labbe2020cosypose,Wang_2021_GDRN} have recently achieved promising performance on benchmark datasets \cite{hodan2020bop}.

Most learning-based approaches \cite{di2021so,he2021ffb6d,labbe2020cosypose,su2022zebrapose,wang2019densefusion} primarily benefit from large amounts of real training images with ground-truth 6D pose labels. 
Nevertheless, annotating the 6D object pose requires expert knowledge, making the pose label acquisition prohibitively expensive and time-consuming for many applications.
As a result, leveraging cost-free auto-annotated synthetic data for training has become indispensable in many recent 6D object pose estimation works \cite{cai2022ove6d,labbe2020cosypose,sundermeyer2020multi}. Yet, these models trained using synthetic data suffer noticeable performance degradation due to the synthetic-to-real domain gap. 

To bridge the domain gap, one seminal work called Self6D \cite{wang2020self6d} was proposed by Wang \etal for self-supervised monocular 6D object pose estimation by leveraging synthetic RGB images for pre-training and then real unlabeled RGB-D images for fine-tuning. Later, Wang \etal presented an extended version called Self6D++ \cite{wang2021occlusion} to better handle object occlusion. Both methods employ the differentiable rendering technique \cite{chen2019learning} to establish the supervision for fine-tuning. Nevertheless, online differentiable rendering introduces an additional burden of computation during training.

\begin{figure}[!t]
\centering
\includegraphics[width=0.95\textwidth]{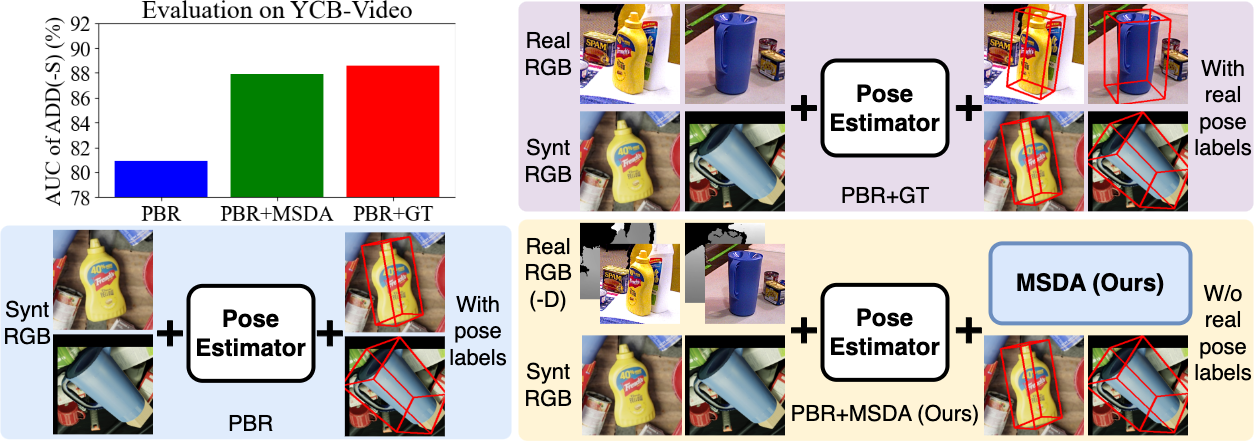}
\caption{Comparison of different training strategies and their corresponding performance on YCB-Video. Top left: performance comparison in terms of the AUC of ADD-(S). Bottom left: the object pose estimator is pre-trained using synthetic physically-based rendering (PBR) images. Top right: the pose estimator is fine-tuned using synthetic and real images with ground-truth real pose labels. Bottom right: the pose estimator is fine-tuned using synthetic and real images based on the proposed MSDA approach without needing real pose labels.}
\label{fig:intro}
\end{figure}

Similarly, this work aims to bridge the synthetic-to-real domain gap for monocular 6D object pose estimation without using real 6D pose labels. To this end, we propose a monocular self-supervised domain adaptation method based on the recent symmetry-agnostic and correspondence-free 6D object pose estimator SC6D \cite{cai2022sc6d}. To supervise the domain adaptation, we first introduce a pose-aware data augmentation based on monocular real RGB images for learning the object pose-aware consistency. Next, we present an effective depth-guided pseudo-label generation strategy to supervise the object distance estimation task. Unlike the state-of-the-art method Self6D++, our method requires neither the object symmetry priors nor the time-consuming online differentiable rendering during training. 

We experimentally evaluate our method on the YCB-Video dataset \cite{xiang2018posecnn} and demonstrate that it outperforms not only the top-tier self-supervised approach Self6D++ but also some state-of-the-art fully-supervised methods using real pose labels. In addition, the experimental results show that our method achieves competitive performance with its fully-supervised counterpart (see Fig. \ref{fig:intro}), indicating our approach is capable of nearly closing the synthetic-to-real domain gap without using real pose labels.

To summarize, our contributions are as follows. 1). We present a pose-aware consistency learning method based on only monocular RGB images for self-supervised domain adaptation. 2). We propose a practical depth-guided pseudo-label generation approach for supervising the object translation estimation along the z-axis. 3). We conduct experiments on the YCB-Video benchmark and demonstrate that our approach outperforms the state-of-the-art self-supervised method Self6D++ and some current fully-supervised methods.

\section{Related Work}
In this section, we mainly focus on recent learning-based 6D object pose estimation approaches that are most related to this work.
\paragraph{Correspondence-based methods.}
This line of work aims to first learn the 2D-3D correspondences, between the 2D image pixel coordinates and the 3D surface coordinates of the object CAD model, from the monocular RGB images and then recover the 6D object poses by solving the Perspective-n-Point (PnP) \cite{lepetit2009epnp} or PnP-RANSAC \cite{fischler1981random} algorithms. BB8 \cite{rad2017bb8} is one of the most representative correspondence-based works and aims to detect the 2D projected pixel locations of the 3D object bounding box corners (keypoints) to build sparse 2D-3D correspondences. Different from BB8, PVNet \cite{peng2020pvnet} selects the pre-defined 3D keypoints from the surface of the 3D object CAD model and then localizes the 2D pixel coordinates of these 3D keypoints in RGB images based on the pixel-wise voting schema. Apart from building these sparse keypoint-based correspondences, learning the dense pixel-wise correspondences \cite{Park_2019_ICCV,li2019cdpn,zakharov2019dpod,hodan2020epos,haugaard2022surfemb} has become mainstream in this field. Pix2Pose \cite{Park_2019_ICCV} directly regresses the dense pixel-wise 2D-3D correspondences from RGB images. Instead, DPOD \cite{zakharov2019dpod} employs a classification-based framework to estimate the 3D coordinates for the 2D object pixels. To better train the object 3D coordinates classification task, ZebraPose \cite{su2022zebrapose} presents an efficient 3D surface coordinate representation using hierarchical binary encoding codes and achieves state-of-the-art performance on the YCB-Video dataset \cite{xiang2018posecnn}. However, most of these methods require prior information about object symmetries to explicitly handle the visual ambiguities. In contrast, EPOS \cite{hodan2020epos} and SurfEmb \cite{haugaard2022surfemb} aim to learn the 2D-3D correspondence distribution over the object surface and can handle visual ambiguity implicitly without needing object symmetry priors.

\paragraph{Regression-based methods.} Instead of recovering the object pose from the intermediate 2D-3D correspondences, the regression-based methods estimate the pose parameters directly from the monocular RGB images in an end-to-end fashion. PoseCNN \cite{xiang2018posecnn} disentangles the 6D pose representation into two components, \ie, the 3D rotation and 3D translation, and separately estimates each component from a monocular RGB image. GDR-Net \cite{Wang_2021_GDRN} proposes a geometry-guided regression network for predicting the 6D pose parameters directly from the estimated 2D-3D correspondences. Based on GDR-Net, SO-Pose \cite{di2021so} further boosts the performance by imposing additional geometry information regarding the object self-occlusion during training. Meanwhile, CosyPose \cite{labbe2020cosypose} and RePose \cite{iwase2021repose} are capable of predicting the relative pose parameters based on an initial pose estimate to refine the initial pose. Still, these approaches are sensitive to visual ambiguities and require additional effort to deal with object symmetries during training, while SC6D \cite{cai2022sc6d} introduces a symmetry-agnostic 6D object pose estimation framework and can effectively deal with visual ambiguities. Therefore, we adapt SC6D as our pose estimator to demonstrate the proposed self-supervised domain adaptation approach. 

\paragraph{Self-supervised methods.} This type of work aims at learning deep models by exploiting massive unlabeled real-world data with related proxy tasks. To the best of our knowledge, Wang \etal \cite{wang2020self6d} presented the first self-supervised monocular 6D object pose estimation work called Self6D. They first train a pose regression network fully supervised with synthetic data, then fine-tune the network using unlabeled real RGB-D images. Self6D adapts an advanced differentiable neural rendering technique DIR-B \cite{chen2019learning} to establish the supervision during fine-tuning. The follow-up work Self6D++ \cite{wang2021occlusion} is developed to account for the occlusion-aware object pose estimation by leveraging both visible and amodal object masks. Meanwhile, Self6D++ employs the much stronger GDR-Net \cite{Wang_2021_GDRN} as the pose estimator, thus outperforming Self6D by a large margin.

\section{Method}

Given an RGB image with the known object, the task is to estimate the 6D object pose $[R|t]$, \ie, the 3D rotation $R \in SO(3)$ and the 3D translation $t \in \mathbb{R}^3$, with respect to the camera coordinate system. We propose a monocular self-supervised domain adaptation (MSDA) approach for the pose estimation task lacking real pose labels. As illustrated in Fig. \ref{fig:intro}, we first pre-train the pose estimator fully supervised with synthetic RGB images generated with the physically-based rendering (PBR) technique \cite{denninger2020blenderproc} and then fine-tune the pose estimator using a mixture of synthetic and real-world data without needing real pose labels. Our method can be easily applied as a plugin on the two-stage based object pose estimators (\ie, first performing the object detection and then estimating the 6D object pose parameters), such as GDR-Net \cite{Wang_2021_GDRN} and SC6D \cite{cai2022sc6d}. In this work, we implement our MSDA approach based on the recent pose estimator SC6D due to its top-performing performance.

\subsection{Framework}
\begin{figure}[!t]
\centerline{
\includegraphics[width=0.99\linewidth]{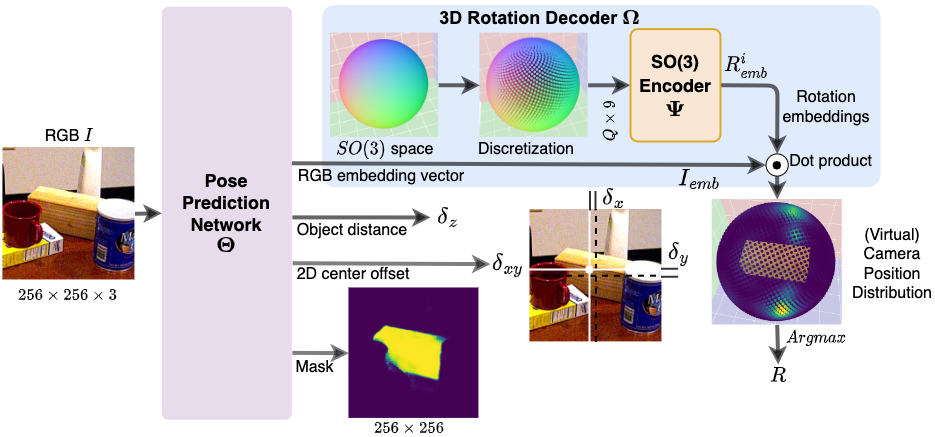}}
\caption{
Overview of SC6D\textsuperscript{++}. 
The pose prediction network $\mathbf{\Theta}$ takes a known object RGB image $I$ as input that predicts the 2D object center offset $\delta_{xy}$, the object distance $\delta_z$, and the image embedding vector $I_{emb}$. Subsequently, the 3D rotation decoder takes as input the embedding vector $I_{emb}$ and outputs the 3D object orientation $R$ with respect to the object itself.  To visualize the estimated 3D rotation probability distribution (see the right part), we highlight the possible positions of the (virtual) camera on a sphere centered upon the object. The brighter color indicates higher probabilities. 
}
\label{fig:inference}
\end{figure}

\paragraph{Architecture.} We build our method upon the recent symmetry-agnostic pose estimator SC6D \cite{cai2022sc6d} that directly predicts the object pose parameters from an object-centric RGB image. We slightly modify the original SC6D to output the same size segmentation mask as the input and denote it as SC6D\textsuperscript{++}, see Fig. \ref{fig:inference}. SC6D\textsuperscript{++} consists of two major components, \ie, a pose prediction network $\mathbf{\Theta}$ and a 3D rotation decoder $\mathbf{\Omega}$, and estimates the object 6D pose from the object-centric RGB image, as shown in Fig.\ref{fig:inference}. The pose prediction network $\mathbf{\Theta}$ takes a known object RGB image $I \in \mathbb{R}^{S \times S \times 3}$ as input and outputs the object distance $\delta_z\in \mathbb{R}$, 2D projection center offset $\delta_{xy} \in \mathbb{R}^2$, segmentation mask $M\in \mathbb{R}^{S \times S}$, and RGB embedding vector $I_{emb}\in \mathbb{R}^{C}$. Meanwhile, the 3D rotation decoder outputs the 3D object rotation matrix $R\in \mathbb{R}^{3 \times 3}$ based on the RGB embedding vector $I_{emb}$.

\paragraph{Scale-invariant 3D translation estimation.} The outputs $\delta_{xy}$ and $\delta_z$ of the pose prediction network $\mathbf{\Theta}$ are grounded on the object-centric RGB image $I$, and this image is obtained using the object bounding box centered at $(x_{bbox}, y_{bbox})$ with the size of $(w_{bbox}, h_{bbox})$. Thus, the 3D object translation is recovered by $t = [t_x, t_y, t_z]^T = r \cdot \delta_z \cdot K^{-1} [o_x, o_y, 1]^T$, where $r=S/\max(w_{bbox}, h_{bbox})$ is the rescaling factor, $[o_x, o_y]^T = S\cdot (\delta_{xy} + 0.5)$ is the projected 2D object center location, and $K$ is the corresponding (virtual) camera intrinsic regarding the rescaled object-centric RGB image $I$ with the fixed scale $S$.

\paragraph{Symmetry-agnostic 3D rotation estimation.} The 3D rotation decoder $\mathbf{\Omega}$ aims to learn the 3D object rotation distribution based on the visual representation to implicitly handle the symmetry ambiguities. To this end, an SO(3) encoder $\mathbf{\Psi}$ is utilized to learn a latent 3D rotation representation $R^i_{emb}\in \mathbb{R}^C$ for each 3D rotation $R^i$ uniformly sampled from the SO(3) space, \ie $R^i_{emb}=\mathbf{\Psi}(R^i)$. The 3D object rotation distribution is approximated by
\begin{equation}
\label{eq:rot_approx}
\begin{split}
     p(R^i|I) &\approx \frac{ \exp( I_{emb} \cdot R^i_{emb} )}{\sum_{j=1}^{N_r} \exp(I_{emb} \cdot R^j_{emb})}, \forall i = 1, 2, \dots N_r,
\end{split}
\end{equation}
where $N_r$ is the number of the 3D rotation samples, and $I_{emb}\in \mathbb{R}^C$ denotes the representation vector learned from the object-centric RGB image $I$. By doing so, the object symmetry ambiguities can be implicitly handled, for example, by mapping all symmetric 3D rotations to similar latent representations correlated with the object observation. The 3D rotation $R$ with the highest probability $p(R|I)$ is thus selected as the prediction.


\paragraph{Supervised learning with synthetic data.}
The pose estimator is pre-trained using synthetic data in a fully supervised manner, and the training loss is summarized as
\begin{equation}
\label{eq:syn_loss}
\begin{split}
    \mathcal{L}^{syn} &= \lambda^{syn}_{xy} \mathcal{L}^{syn}_{xy} + \lambda^{syn}_{z} \mathcal{L}^{syn}_{z} + \lambda^{syn}_{R} \mathcal{L}^{syn}_{R} + \lambda^{syn}_{M} \mathcal{L}^{syn}_{M}, \\
\end{split}
\end{equation}
where $\mathcal{L}^{syn}_{\{xy, z, R, M\}}$ represent the training objectives for supervising the estimation tasks of the 2D object center offset $\delta_{xy}$, the object distance $\delta_{z}$, the 3D object rotation $R$, and the object segmentation mask $M$, and $\lambda^{syn}_{\{xy, z, R, M\}}$ denote the  balancing parameters for the corresponding loss terms. We kindly refer the reader to \cite{cai2022sc6d} for more implementation details.

\paragraph{Initial 6D pose estimate.} Like Self6D++ \cite{wang2021occlusion}, our method requires an initial 6D pose estimate from the real RGB image to establish the supervision. Thus, we leverage the pre-trained pose estimator to provide the initial 6D pose $[R|t]$ utilized for our self-supervised domain adaptation.

\subsection{Self-supervised Pose-Aware Consistency Learning}
\begin{figure}[t]\centerline{\includegraphics[width=0.99\linewidth]{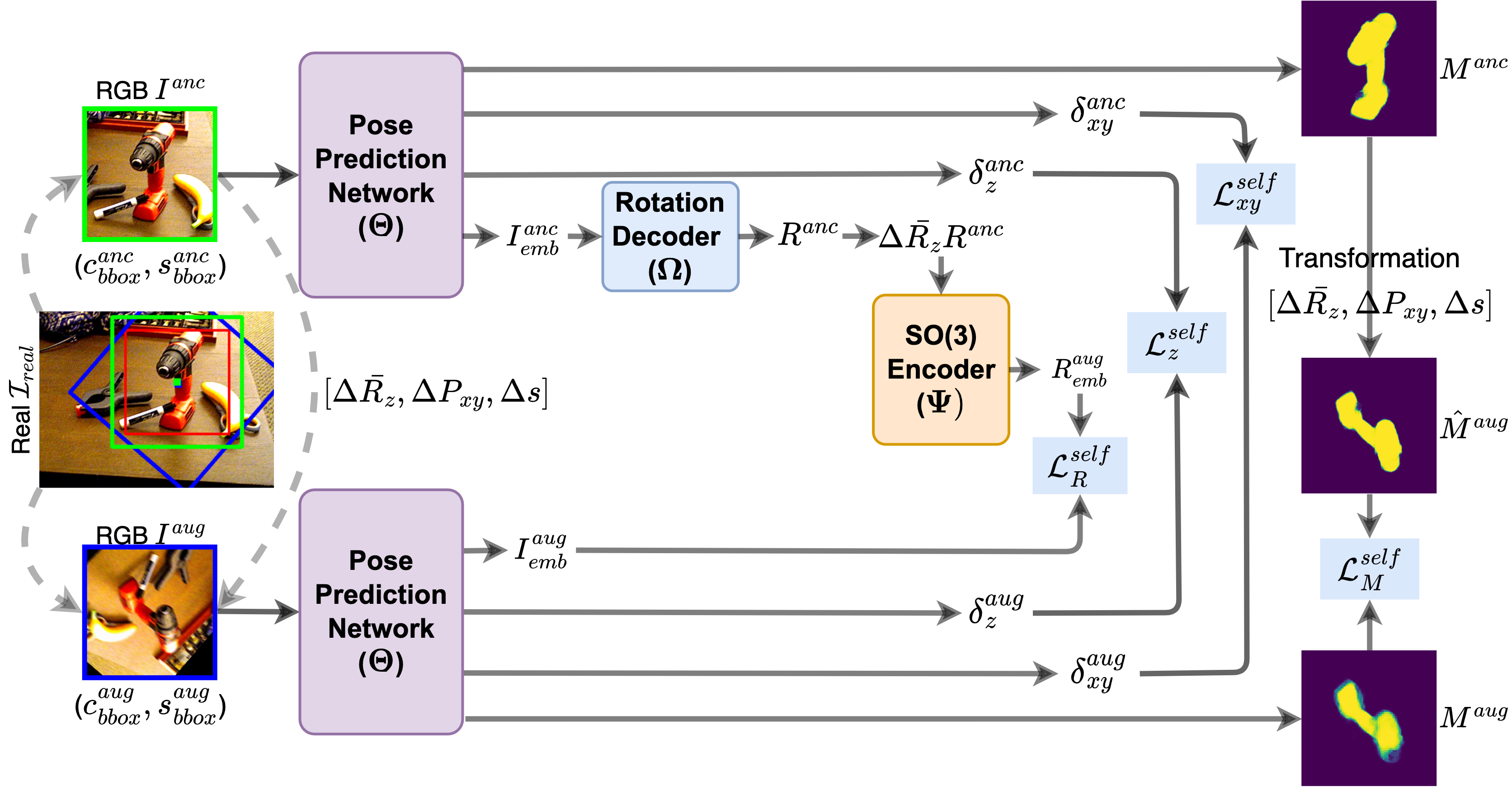}}
\caption{
Illustration of the pose-aware consistency learning for self-supervised domain adaptation. Given a real RGB image $\mathcal{I}_{real}$ and an object bounding box (red box), we first obtain an anchor RGB sample $I^{anc}$ using an expanded box (green box) and an augmented RGB sample $I^{aug}$ using a transformed box (blue box) with a known relative pose $\Delta=[\Delta{\bar{R}_z}, \Delta{P_{xy}}, \Delta{s}]$. Then the pose prediction network $\mathbf{\Theta}$ separately predicts the outputs of the paired samples ($I^{anc}, I^{aug}$),\ie, the object center projection offsets ($\delta^{anc}_{xy}, \delta^{aug}_{xy}$), the object distances ($\delta^{anc}_{z}, \delta^{aug}_{z}$), the RGB embedding vectors ($I^{anc}_{emb}, I^{aug}_{emb}$), and the segmentation masks ($M^{anc}, M^{aug}$). The supervision signals are established by enforcing the pose-aware consistencies between the anchor sample and its augmented one. The network weights in all modules are shared during training.
}
\label{fig:uda_pipe}
\end{figure}

We first introduce our self-supervised pose-aware consistency learning based on monocular RGB images. Given a real RGB image $\mathcal{I}_{real}\in \mathbb{R}^{H \times W \times 3}$ with an object bounding box $B_{real}=(x_{bbox}, y_{bbox}, w_{bbox}, h_{bbox})$, we generate a pair of object-centric RGB samples with the known relative pose and enforce the pose estimator to learn the pose-aware consistency (see Fig.\ref{fig:uda_pipe}).

Specifically, we first obtain an anchor sample $I^{anc}$ using an enlarged square bounding box $B^{anc}$ centered at $c^{anc}_{bbox}=(x_{bbox}, y_{bbox})$ with the scale of $s^{anc}_{bbox}=f_{anc}\cdot\max(w_{bbox}, h_{bbox})$, where $f_{anc}$ is a random scaling factor. Then, we transform the anchor bounding box $B^{anc}$ using a random rescaling factor $\Delta{s}$, a random offset $\Delta{P_{xy}}$, and a random in-plane rotation $\Delta{R_z}\in \mathrm{SO(2)}$ to obtain an augmented variant $I^{aug}$ for the anchor sample dynamically. The pose relationship between the training pair $(I^{aug}, I^{anc})$ can be formulated as 
\begin{equation} \label{eq:aug_pose}
\begin{split}
\begin{cases}
    R^{aug} &= \Delta{\bar{R_z}}R^{anc} \\
    \delta^{aug}_{z} &= \Delta{s} \cdot \delta^{anc}_{z} \\
    \delta^{aug}_{xy} &= \Delta{R_z} (\delta^{anc}_{xy} - \Delta{P_{xy}} / s^{anc}_{bbox}) / \Delta{s} \\
\end{cases},
\end{split}
\end{equation}
where $[\delta^{anc}_{xy}, \delta^{anc}_{z}, R^{anc}]$ and $[\delta^{aug}_{xy}, \delta^{aug}_{z}, R^{aug}]$ are the predictions of the model for the anchor sample $I^{anc}$ and its augmented one $I^{aug}$, respectively. $\Delta{\bar{R_z}} \in \mathbb{R}^{3\times 3}$ denotes the $3\times 3$ matrix representation of $\Delta{R_z}$.

By this means, we construct online pairwise training samples $(I^{aug}, I^{anc})$ with the known relative transformation $\Delta = [\Delta{s}, \Delta{P_{xy}}, \Delta{R_z}]$ based on monocular real RGB images without using pose labels. Subsequently, we leverage the L1 loss to measure the translation consistency, \ie,
\begin{equation} 
\label{eq:self_tsl}
\begin{split}
\begin{cases}
\mathcal{L}^{self}_{z} = ||\delta^{aug}_{z}  -  \Delta{s} \cdot \delta^{anc}_{z} ||_1 \\
\mathcal{L}^{self}_{xy} = ||\delta^{aug}_{xy} - \Delta{R_z} (\delta^{anc}_{xy} - \Delta{P_{xy}} / s^{anc}_{bbox}) / \Delta{s} ||_1 \\
\end{cases}.
\end{split}
\end{equation}
Regarding the rotation consistency, we enforce the network to minimize the negative log-likelihood $p(R^{aug}|I^{aug})$,
\begin{equation} 
\begin{split}
\mathcal{L}^{self}_{R} &= -\log(p(R^{aug}|I^{aug})) \\
&= -\log\frac{\exp(I^{aug}_{emb} \cdot R_{emb}^{aug} / \tau) }{\sum^{N_r}_i \exp(I_{emb}^{aug} \cdot R_{emb}^i / \tau)} \\
\end{split},
\end{equation} 
where $I^{aug}_{emb} \in \mathbb{R}^C$ is the image embedding vector of the augmented sample $I^{aug}$, $R^{aug}_{emb} = \mathbf{\Psi}(R^{aug})$ is the latent representation vector of the derived 3D rotation $R^{aug}$, and $\tau=0.1$ is the temperature parameter.

In addition, we also leverage the object segmentation consistency loss $\mathcal{L}^{self}_{M}$ to supervise the domain adaptation process. To do this, we first transform the 2D pixel coordinate grid $G^{anc}$ of the anchor RGB sample $I^{anc}$ based on the transformation $\Delta$ and then sample the segmentation mask $\hat{M}^{aug}$ from the predicted anchor mask $M^{anc}$ according to the transformed $G^{anc}$ (see the right side in Fig.\ref{fig:uda_pipe}). Therefore, the self-supervised pose-aware consistency loss is summarized as
\begin{equation}
\begin{split}
    \mathcal{L}^{self} = \lambda_{xy}^{self} \mathcal{L}^{self}_{xy} + \lambda^{self}_{z} \mathcal{L}^{self}_{z} + \lambda^{self}_{R} \mathcal{L}^{self}_{R} + \lambda^{self}_{M} \mathcal{L}^{self}_{M} \\
\end{split}
\end{equation}
where $\lambda^{self}_{\{xy, z, R, M\}}$ denote the weight hyper-parameters for the corresponding loss terms.

\subsection{Depth-guided Object Distance Learning}

\begin{figure}[!t]
\centerline{
\includegraphics[width=0.99\linewidth]{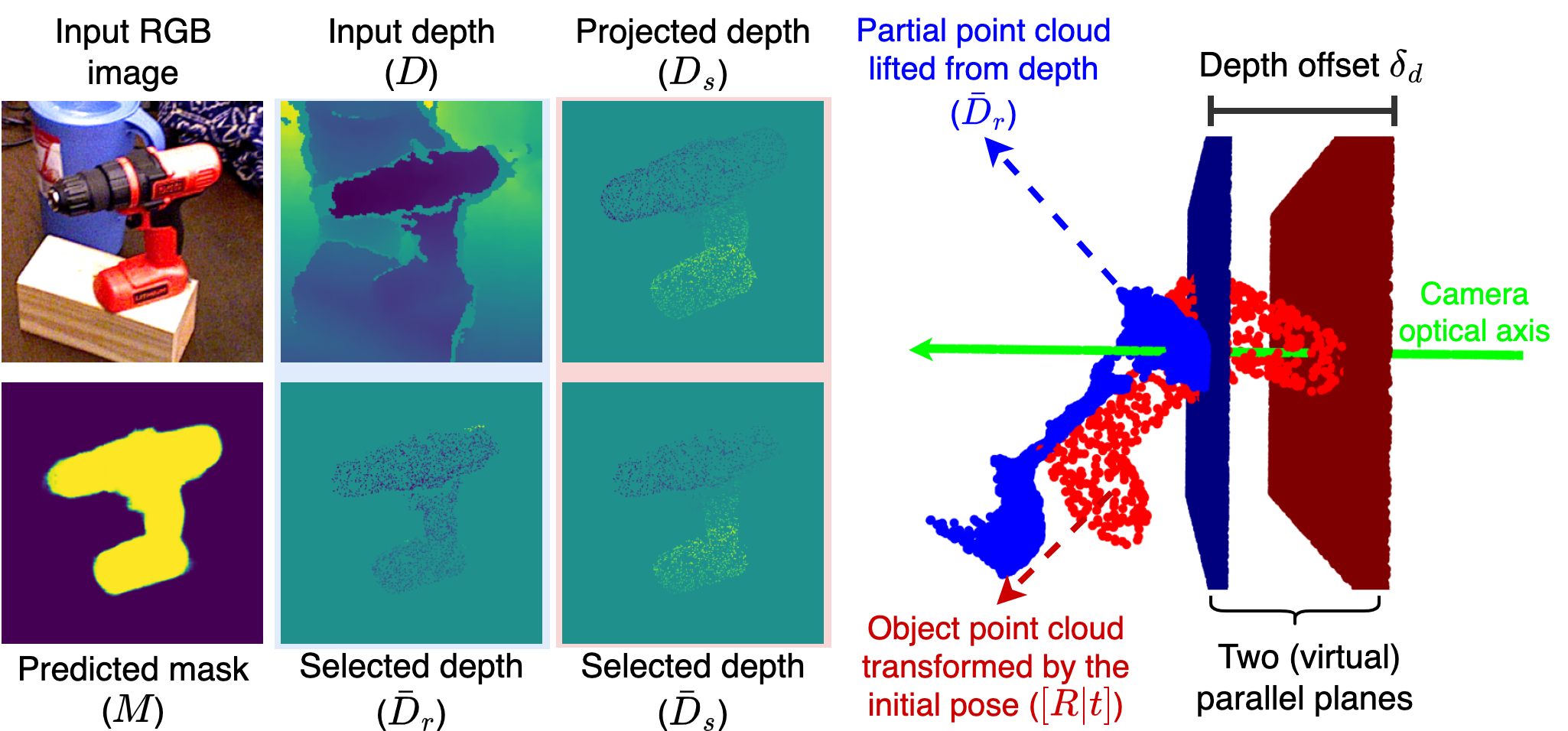}}
\caption{
Illustration of the depth-guided object z-axis translation pseudo-label generation. Given the object-centric RGB and depth images, we first utilize the pre-trained pose estimator to predict the object mask $M$ and an initial pose $[R|t]$ from the RGB image. Then we transform the 3D object point cloud using the estimated initial pose and the corresponding camera intrinsic to obtain the projected depth image $D_s$. A depth set $\bar{D}_s$ is sampled from the projected depth $D_s$ based on the predicted mask $M$, and another depth set $\bar{D}_r$ is sampled from the input depth image $D$ based on both $M$ and $D_s$. On the right side, the depth offset $\delta_d$ between two point clouds is considered as the distance between two virtual parallel planes perpendicular to the camera's optical axis. One plane intersects with the closest point in the partial point cloud (\textcolor{blue}{blue}) lifted from the observed depth image $\bar{D}_r$. The other plane intersects with the closest point in the complete 3D object point cloud (\textcolor{red}{red}) transformed by the initial pose estimate $[R|t]$.
}
\label{fig:depth}
\end{figure}

When depth images $\mathcal{D}_{real}$ are available, we propose a depth-guided pseudo-label generation method and utilize the generated \textit{pseudo-label} $\bar{t}_z$ to supervise the object translation estimation along the z-axis ($t_z$). However, it is challenging to estimate the object distance (\ie, from the camera origin to the object centroid) from the depth image directly because these depth values represent the measurements from the camera plane to the surface of the 3D object. Moreover, this task becomes more challenging when the object is occluded and only partially visible in the depth image. 

To tackle these issues, we propose estimating a depth offset $\delta_d$ between the two closest points, one from the observed real depth image and the other from the 3D object point cloud transformed using the initial pose $[R|t]$, to compensate for the initial z-axis translation estimate $t_z$. To this end, we first obtain the object-centric real depth image $D \in \mathbb{R}^{S\times S}$ using the anchor bounding box $B^{anc}$, then the 3D object points $P_s \in \mathbb{R}^{N_s \times 3}$ are uniformly sampled from the surface of the 3D object model, where $N_s$ is the number of the 3D points. Next, we transform the sampled 3D object points $P_s$ using the estimated pose $[R|t]$ and the intrinsic $K$, \ie,
\begin{equation}
    D^{i}_{s}[U_{s}^i, V_{s}^i, 1]^T =K (R P_s^i+ t), ~~~ \forall i=1, \dots, N_s \\
    \label{eq:proj}
\end{equation}
where $D^{i}_{s}$ and $(U_{s}^i, V_{s}^i)$ denote the depth value and the projected 2D pixel coordinate of the 3D object point $P_s^i$ after the transformation, respectively.

Based on the predicted object mask $M$ and the projected coordinate $(U_s^i, V_s^i)$, we select the corresponding depth values from the set $\{D^{i}_{s}\}_{i=1}^{N_s}$ and the object-centric real depth image $D$. This process can be formulated as
\begin{equation}
\begin{split}
\bar{D}^{i}_s &= 
    \begin{cases}
        D_s^i, \textit{ if } \{U^i_s, V_s^i\} \in M ~\textit{ and }~ M_{U^i_s, V_s^i} \geq \rho, \\ 
        \epsilon, ~~~~~\textit{otherwise},  \\
    \end{cases} \\
\bar{D}^{i}_r &= 
    \begin{cases}
        D_{U^i_s, V_s^i}, \textit{ if } \{U^i_s, V_s^i\} \in M \textit{ and } M_{U^i_s, V_s^i} \geq \rho, \\ 
        \epsilon, ~~~~~~~~\textit{otherwise},  \\
    \end{cases}
\end{split}
\end{equation}
where $\bar{D}^i_s$ and $\bar{D}^{i}_r$ are the sampled depth values from the transformed 3D object point cloud and the observed depth image $D$ (see the left side in Fig. \ref{fig:depth}), respectively. $\rho=0.9$ is the confidence threshold, and $\epsilon \gg \max(D) $ denotes the invalid (penalty) depth value for the background.

In this way, we obtain two depth sets, \ie, $\bar{D}_s = \{\bar{D}^{i}_s\}^{N_s}_{i=1}$ and $\bar{D}_r = \{\bar{D}^{i}_r\}^{N_s}_{i=1}$, the former from the transformed 3D object point cloud $P_s$ and the latter from the observed depth image $D$. The depth offset $\delta_d$ between these two sets can be computed to generate the \textit{pseudo-label} $\bar{t}_z$ for the object z-axis translation, \ie, $\bar{t}_z = \delta_d + t_z$. However, it is non-trivial to compute this offset $\delta_d$ because $\bar{D}_s$ is derived from the entire 3D object point cloud while $\bar{D}_r$ is sampled from its partial observation. To approximate $\delta_d$, we compute the depth offset between the closest point observed in $\bar{D}_r$ and the one from $\bar{D}_s$, \ie, $\delta_d = \min (\bar{D}_r) - \min (\bar{D}_s)$.
In other words, $\delta_d$ can be regarded as the distance between two planes perpendicular to the camera optical axis, where one plane intersects with the closest point in the observed object depth image, and the other intersects with the closest point in the transformed object point cloud, see the right in Fig. \ref{fig:depth}.

However, outliers could be included in the real depth set $\bar{D}_r$ and result in significant errors due to the inaccurate mask prediction $M$. Instead of simply selecting the closest real depth $\bar{D}^{o}_r = \min (\bar{D}_r)$, we propose an adaptive selection strategy to compute $\bar{D}^{o}_r$. We first rank the depth set $\bar{D}_r$ in ascending order, followed by a moving average operation, to obtain a smooth sorted set $\hat{D}_r$ and then calculate the adjacent depth difference $\hat{G}_r$ of the sorted set. Finally, the real depth value $\bar{D}^{o}_r = \hat{D}_r^k$ is selected if the first $k^{th} \in [1, N_s-1]$ difference value in $\hat{G}_r$ is smaller than the predefined threshold, \ie $\hat{G}_r^k \leq \gamma$ where $\gamma=0.001$ in our experiments, otherwise $\bar{D}^{o}_r = \min (\bar{D}_r)$. We utilize the generated \textit{pseudo-label} $\bar{t}_z$ and a truncated L1 loss to supervise the training process, 
\begin{equation}
    \begin{split}
        \mathcal{L}^{pseudo}_{tz} = \min(||\bar{t}_z - t_z||_1, \xi)
    \end{split}
\end{equation}
where $\xi=0.1$ is an upper-bound value for the distance offset to avoid the unstable loss.

\subsection{Joint Fine-tuning for Domain Adaptation}
We employ both synthetic and real data to jointly fine-tune the pre-trained pose estimator. To summarize, the overall training objective is written as
\begin{equation}
\begin{split}
    \mathcal{L} &= \lambda^{syn} \mathcal{L}^{syn} +  \lambda^{self} \mathcal{L}^{self} + \lambda^{pseudo}_{tz} \mathcal{L}^{pseudo}_{tz},
\end{split}
\end{equation}
where $\lambda^{syn}, \lambda^{self}$, and $\lambda^{pseudo}_{tz}$ represent the weight hyper-parameters of the corresponding loss terms.

\section{Experiment}

\paragraph{\textbf{Dataset.}} The experiments are conducted on the benchmark YCB-Video \cite{xiang2018posecnn} dataset. YCB-Video \cite{xiang2018posecnn} contains 21 objects with textured 3D CAD models and 92 annotated real RGB-D videos collected from multiple noisy environments. More than 113,000 images from 80 videos are used as the training data, and 2949 keyframes are selected from the rest 12 videos as the testing set. The physically-based rendering (PBR) dataset \cite{hodan2020bop} contains 50,000 synthetic images and is employed for training in this work.

\paragraph{\textbf{Evaluation Metrics.}} 
We report our results in terms of the standard metrics of ADD(-S) \cite{hinterstoisser2012model} with the error tolerance threshold less than the 10\% diameter of the object and the AUC (area under the curve) of ADD(-S)/ADD-S 
\cite{xiang2018posecnn} with varying tolerance thresholds up to 10cm to fairly compare with other approaches. We kindly refer the reader to \cite{hinterstoisser2012model,xiang2018posecnn} for the definition details.

\paragraph{\textbf{Implementation Details.}} We utilize the AdamW solver \cite{loshchilov2017decoupled} with the cosine annealing learning rate to train the model on 16 Nvidia GPUs. The pose estimator is first pre-trained using synthetic data for 75 epochs with the learning rate starting from $5\times 10^{-4}$ to $1\times 10^{-5}$ and then fine-tuned using a mixture of synthetic and real data for additional five epochs with the learning rate from $1\times 10^{-4}$ to $1\times 10^{-5}$. We set the weight hyper-parameters $\lambda^{syn}_{xy}=10.0, \lambda^{syn}_{z}=1.0, \lambda^{syn}_{R}=1.0, \lambda^{syn}_{M}=10.0, \lambda^{self}_{xy}=10.0, \lambda^{self}_{z}=10.0, \lambda^{self}_{R}=0.1, \lambda^{self}_{M}=10.0, \lambda^{syn}=1.0, \lambda^{self}=0.1, \lambda^{pseudo}_{tz}=10.0$ to maintain comparable magnitudes among all loss terms. 
Besides, we follow the 3D rotation sampling strategy in \cite{park2020latentfusion,cai2022ove6d} and uniformly sample $N_r=5000$ rotations for training and $N_r=480,000$ ($4000$ viewpoints on a sphere with $120$ in-plane rotations for each viewpoint) for testing. We apply strong image augmentations on synthetic RGB images following \cite{cai2022sc6d} and dynamically crop the region of interest using the in-plane rotated object bounding box during training. For evaluation, we leverage the predicted object bounding boxes provided by Self6D++ \cite{wang2021occlusion} for a fair comparison, where the object detector is trained using purely synthetic PBR images.

\begin{table}[!t]
\caption{Ablation studies using the YCB-Video benchmark in terms of Average Recall (\%) of ADD(-S) metric, and AUC of ADD-S/ADD(-S) metrics. "PBR" indicates the synthetic PBR training images. "LB" denotes the lower bound result only using PBR for training, and "UB" represents the upper bound result obtained using both PBR and real data for training with ground-truth pose labels.
}
\label{tab:ablate_ycbv}
\centering
\renewcommand{\arraystretch}{1.2}
\small
\begin{tabular}{ c | c | l c c c}
\hline
\shortstack{~ \\ ~  \\ ~} 
& \shortstack{~ \\ Training Data \\ ~} 
& \shortstack{~ \\ ~~~~~~~~~~~~Method \\ ~} 
& \shortstack{~ \\ ADD(-S) \\ ~ }  
& \shortstack{~ \\ AUC of \\ ADD-S }  
& \shortstack{~ \\ AUC of \\ ADD(-S)}   \\
\hline
A1 &~ PBR         &~  $\textrm{SC6D}^{++}$(LB)    & 56.1 & 87.5 & 80.9 \\
A2 ~&~ PBR + Real (RGB) ~&~ $\textrm{SC6D}^{++}$(UB)   & \textit{79.8} & \textit{92.2} & \textit{88.6} \\ %
\hline
B1 &~ PBR + Real (RGB) ~&~  $\textrm{SC6D}_{\textrm{MSDA}}^{++}$ w/o $\mathcal{L}^{pseudo}_{tz}$ & 59.4 & 88.4 & 82.2 \\
B2 &~ PBR + Real (RGB-D) ~&~  $\textrm{SC6D}_{\textrm{MSDA}}^{++}$ w/o $\mathcal{L}^{self}$      & 77.6 & 91.6 & 87.2   \\ 
B3 &~ PBR + Real (RGB-D)  ~&~  $\textrm{SC6D}_{\textrm{MSDA}}^{++}$(Ours) & \textbf{80.7} & \textbf{92.1} & \textbf{87.9} \\
\end{tabular}
\end{table}

\subsection{Ablation Study}
We first conduct ablation studies using the proposed self-supervised domain adaptation approach based on monocular RGB(-D) images. The experimental results are reported in Table \ref{tab:ablate_ycbv} in terms of the average recall of ADD(-S) and the AUC of ADD-S/ADD(-S).

Compared to the baseline (A1) fully supervised with synthetic data only (A1), using all the proposed self-supervised training losses (B3) can significantly boost the performance from 56.1\% to 80.7\% in terms of ADD(-S) and from 80.9\% to 87.9\% in terms of AUC of ADD(-S). Specifically,  $\textrm{SC6D}_{\textrm{MSDA}}^{++}$ \textit{w/o} $\mathcal{L}^{pseudo}_{tz}$ improves the ADD(-S) recall by over 3\% using only the pose-aware consistency loss based on monocular real RGB images (B1). In addition, $\textrm{SC6D}_{\textrm{MSDA}}^{++}$ \textit{w/o} $\mathcal{L}^{self}$ can achieve 77.6\% ADD(-S) recall using the loss only based on the object depth-guided z-axis translation \textit{pseudo-labels} (B2). 

Overall, our $\textrm{SC6D}_{\textrm{MSDA}}^{++}$ achieves optimal performance using both the object RGB-based pose-aware consistency and the object depth-guided \textit{pseudo-label} distance losses (B3). Noteworthy, $\textrm{SC6D}_{\textrm{MSDA}}^{++}$ consistently improves over its baseline $\textrm{SC6D}^{++}$(LB) and approaches towards its fully-supervised counterpart $\textrm{SC6D}^{++}$(UB), even slightly outperforming its fully-supervised counterpart regarding the ADD(-S) metric (80.7\% \textit{vs.} 79.8\%), which indicates $\textrm{SC6D}_{\textrm{MSDA}}^{++}$ can nearly bridge the synthetic-to-real domain gap.

\begin{table}[!t]
\caption{Comparison with state-of-the-art methods in terms of Average Recall (\%) of ADD(-S), and AUC of ADD-S/ADD(-S). "Synt + Real GT" denotes using both synthetic and real data with ground-truth pose labels for fully-supervised training. "Synt + Real Self" denotes using synthetic and real data for self-supervised training without needing real pose labels. "UB" indicates the upper bound results obtained using real pose labels. The best results are highlighted in \textbf{bold}, and the second-best results are underlined.  "-" indicates unavailable results.
}
\label{tab:eval_ycbv}
\centering
\renewcommand{\arraystretch}{1.0}
\small
\begin{tabular}{ c | l c c c}
\hline
\shortstack{~ \\ Training \\ ~} 
& \shortstack{~ \\ ~~~~~~~ Method \\ ~} 
& \shortstack{~ \\ ADD(-S) \\ ~ } 
& \shortstack{~ \\ AUC of \\ ADD-S }  
& \shortstack{~ \\ AUC of \\ ADD(-S)}   \\
\hline
\multirow{11}{*}{\shortstack{ 
~\\ Supervision: \\
Synt GT + Real GT
\\ ~}}
 ~&~ PoseCNN \cite{xiang2018posecnn}         & 21.3 & 75.9 & 61.3 \\
 ~&~ PVNet \cite{peng2020pvnet}              & - & - & 73.4       \\
 ~&~ Seg-Driven \cite{hu2019segmentation}    & 39.0 & - & -       \\
 ~&~ SingleStage \cite{hu2020single}         & 53.9 & - & -       \\
 ~&~ CosyPose \cite{labbe2020cosypose}       & - & 89.8 & 84.5    \\
 ~&~ SO-Pose \cite{di2021so}                 & 56.8 & 90.9 & 83.9 \\
 ~&~ GDR-Net \cite{Wang_2021_GDRN}           & 60.1 & 91.6 & 84.4 \\
 ~&~ RePose \cite{iwase2021repose}           & 62.1 & 88.5 & 82.0 \\
 ~&~ ZebraPose \cite{su2022zebrapose}        & \underline{80.5} & 90.1 & 85.3 \\
 ~&~ Self6D++(UB)\cite{wang2021occlusion}    &   -  & 90.7 & 82.6 \\
 ~&~ SC6D \cite{cai2022sc6d}                 & 77.1 & 90.8 & 86.4 \\
 ~&~  $\textrm{SC6D}^{++}$ (UB)               & 79.8 & \textbf{92.2} & \textbf{88.6}  \\ 
\hline
\multirow{2}{*}{\shortstack{ Supervision: \\ Synt GT + Real Self }}
 ~&~ Self6D++ \cite{wang2021occlusion} & - & 91.1 & 80.0    \\
 ~&~  $\textrm{SC6D}_{\textrm{MSDA}}^{++}$ (Ours)  & \textbf{80.7} & \underline{92.1} & \underline{87.9} \\ 
\end{tabular}
\end{table}

\subsection{Comparison with State-of-the-art}
In this part, we compare $\textrm{SC6D}_{\textrm{MSDA}}^{++}$ against current state-of-the-art 6D object pose estimation approaches on the YCB-Video dataset. The comparison results are presented in Table \ref{tab:eval_ycbv} in terms of the standard metrics, ADD(-S) and AUC of ADD-S/ADD(-S). In general, $\textrm{SC6D}_{\textrm{MSDA}}^{++}$ achieves state-of-the-art performance regarding the ADD(-S) metric and outperforms the top-tier self-supervised approach Self6D++ \cite{wang2021occlusion} in terms of the AUC of ADD-S/ADD(-S).

On the one hand, our self-supervised $\textrm{SC6D}_{\textrm{MSDA}}^{++}$ achieves superior performance over the state-of-the-art fully-supervised pose estimators ZebraPose \cite{su2022zebrapose} and SC6D \cite{cai2022sc6d}. Concretely, ZebraPose \cite{su2022zebrapose} is a top-performing correspondence-based pose estimator achieving 80.5\% ADD(-S) recall using real pose labels for training, in comparison, $\textrm{SC6D}_{\textrm{MSDA}}^{++}$ can reach 80.7\% without needing real pose labels. Meanwhile, compared with the original symmetry-agnostic SC6D \cite{cai2022sc6d}, $\textrm{SC6D}_{\textrm{MSDA}}^{++}$ outperforms it by 3.6\% and 1.6\% regarding the metrics of ADD(-S) and AUC of ADD(-S) despite the lack of real pose labels, correspondingly. On the other hand, $\textrm{SC6D}_{\textrm{MSDA}}^{++}$ also outperforms the current state-of-the-art self-supervised pose estimator, Self6D++ \cite{wang2021occlusion}. In particular, without using real pose labels, Self6D++ achieves 91.1\%/80.0\% AUC of ADD-S/ADD(-S) behind 92.1\%/87.9\% obtained by $\textrm{SC6D}_{\textrm{MSDA}}^{++}$. 

\section{Conclusion}

This work introduces a monocular self-supervised domain adaptation approach for 6D object pose estimation. By leveraging real-world RGB(-D) images, our method is capable of bridging the synthetic-to-real domain gap and achieving superior performance on the YCB-Video benchmark without using real pose labels. We accomplish this by learning the object pose-aware consistency based on real-world monocular RGB images and compensating for the inaccurate object translation estimation along the z-axis using the observed depth images. Our approach does not rely on complex differentiable rendering techniques to establish supervision, simplifying the self-supervised domain adaptation process for the 6D object pose estimation task.

\section{Acknowledgement}
This work was supported by the Academy of Finland under the project \#327910.

\bibliographystyle{splncs04}
\bibliography{pose}
\end{document}